\documentclass[conference]{IEEEtran}
\IEEEoverridecommandlockouts

\usepackage{cite}
\usepackage{amsmath,amssymb,amsfonts}
\usepackage{algorithmic}
\usepackage{graphicx}
\usepackage{textcomp}
\usepackage{xcolor}
\usepackage{listings}
\usepackage{booktabs}
\usepackage{url}
\usepackage{array}
\usepackage[hidelinks]{hyperref}
\usepackage{makecell}

\begin{document}

\title{Training-Free Occluded Text Rendering via Glyph Priors and Attention-Guided Semantic Blending}

\author{\IEEEauthorblockN{Jingqi Hou$^{1}$, Hongtian Wang$^{2}$}
\IEEEauthorblockA{$^{1,2}$College of Computer Science, Beijing University of Technology, Beijing, China\\
$^{1}$Email: lcstate@emails.bjut.edu.cn\\
$^{2}$Email: laowang@emails.bjut.edu.cn}
}

\maketitle

\begin{abstract}
We present a training-free framework for occluded text rendering with a pretrained FLUX.1-dev backbone. The task requires a model to render recognizable typography and place an occluding object over the intended text region. This setting remains difficult for existing text-to-image generators: the occluder often drifts away from the text, while the text may be distorted or appear to float on top of the occluding object. To address this problem, we propose a restarted dual-stream inference framework that decouples text-layout preservation from occluder insertion. A Base Stream provides a clean typographic reference and same-step key/value (K/V) features, while the Edit Stream is conditioned on the occlusion prompt. We further adopt the spectral glyph-prior idea from FreeText and adapt it to stabilize the target text structure during early-to-mid denoising. In the reasoning pass, our method localizes the target text, estimates a text-band region from token-conditioned attention and glyph support, and derives an anchor-aware hard fusion mask for the occluder. In the final edit pass, generation restarts from the same initial noise and applies hard mask-guided image-token K/V replacement at selected attention sites, preserving the Base layout outside the mask while injecting the occluder appearance from the Edit Stream inside the mask. Experiments on representative occluded text scenarios demonstrate substantially improved text readability and competitive occlusion alignment, yielding more stable object-on-text compositions without any model fine-tuning.
\end{abstract}


\section{Introduction}

Visual text is not merely a decorative pattern in images, but a functional carrier of semantic information. In practical scenarios such as posters, product packaging, advertising creatives, storefront signs, and user interfaces, the generated text must remain both visually plausible and semantically readable. Minor spelling errors, malformed glyphs, or broken layout can already make an otherwise appealing image unusable. For this reason, recent studies increasingly view text rendering as a stringent test of fine-grained controllability and cross-modal alignment in text-to-image generation \cite{freetext,textcrafter}.

Despite the remarkable progress of modern text-to-image models, reliable visual text generation remains difficult, especially in multi-line, text-dense, and semantically complex scenes \cite{freetext,textcrafter}. The problem becomes even harder in the setting studied in this paper: \emph{occluded text rendering}. Here, the model must not only render recognizable typography, but also place a newly generated object over the intended text region while preserving the surrounding text layout. This requires the model to satisfy two conflicting goals simultaneously: preserving the structural regularity of text and enforcing a plausible front--back relationship between text and object.

Unfortunately, existing approaches do not fully address this setting. On the one hand, prior visual text generation methods mainly focus on improving the rendering accuracy and controllability of text itself, rather than handling local object-on-text occlusion \cite{textdiffuser,anytext,textdiffuser2,freetext}. Many of them rely on retraining, external glyph/layout supervision, or rigid positional conditioning, which can limit flexibility and may interfere with the base model's native scene planning \cite{freetext}. On the other hand, occlusion-aware generation methods aim to control front--back relationships between objects, but are not designed to preserve fine-grained typography under local overlap \cite{larender}. Inpainting- or mask-based local editing pipelines often require user-specified masks and may produce visually hard boundaries or imperfect integration when the edited region must remain structurally coherent \cite{bld}. Attention-control and feature-sharing methods can improve editing consistency, but they are not inherently tailored to text-preserving localized occlusion, where small spatial errors can noticeably degrade text readability \cite{ptp,freeflux}.

These limitations are particularly evident on strong FLUX-like backbones. In our preliminary experiments with FLUX.1-dev, a single occlusion prompt can often generate both the target text and the occluding object, but fails to establish the intended object-on-text relationship: the object may drift away from the text, or the text may remain unnaturally unobstructed. This suggests that prompt semantics alone are insufficient for precise text-centric occlusion control. At the same time, recent work on rotary position embedding (RoPE)-based multimodal diffusion transformer (MMDiT) models has shown that training-free editing on FLUX should not rely on uniform intervention, but instead benefit from task-specific layer selection and reasoning-aware control \cite{freeflux}.

To address this problem, we present a training-free framework for occluded text rendering on a pretrained FLUX.1-dev backbone. Our method decouples the task into two coordinated objectives: preserving a clean typographic layout and inserting an occluding object into a controlled subregion of the text. At a high level, we combine glyph-prior regularization, dual-stream reasoning, and hard mask-guided image-token K/V replacement to achieve text-preserving localized occlusion without model fine-tuning.

The key contributions of our work are as follows:
\begin{itemize}
    \item We formulate \emph{occluded text rendering} as a new inference-time control problem, where a model must simultaneously preserve typographic layout and generate a plausible occluding object in a controlled text region.
    
    \item We propose a training-free FLUX.1-dev framework that integrates glyph-prior regularization, restarted dual-stream reasoning, and hard mask-guided image-token K/V replacement to achieve text-preserving localized occlusion.
    
    \item Quantitative and qualitative results demonstrate that our method achieves the best text readability and competitive occlusion alignment, providing a better balance between readable typography and object-on-text placement without model fine-tuning.
\end{itemize}

\section{Related Work}

\subsection{Text-to-Image Backbones and Inference-Time Control}
Text-to-image generation has evolved from large latent-diffusion backbones such as SDXL to transformer- and flow-based architectures with stronger prompt following and image synthesis quality \cite{sdxl,scalingrf,flux}. SDXL significantly improved open latent diffusion models for high-resolution image synthesis \cite{sdxl}. More recent MMDiT and rectified-flow designs replace the U-Net-style denoiser with transformer-based architectures, enabling stronger interaction between text and image tokens \cite{scalingrf}. In particular, FLUX.1-dev is an open-weight rectified-flow transformer released by Black Forest Labs and has become an important backbone for inference-time customization and editing research \cite{flux}.

This architectural transition is important for our setting, since our method is built on a pretrained FLUX.1-dev backbone and is more closely related to inference-time control on transformer-based text-to-image generators than to retrained task-specific text rendering models.

\subsection{Visual Text Rendering in Image Generation}
A major line of research improves the ability of text-to-image models to render readable and controllable text in images. Early representative methods such as TextDiffuser formulate visual text generation as a layout-then-render process, using a dedicated layout planner and diffusion-based image synthesis conditioned on textual layout \cite{textdiffuser}. AnyText further improves multilingual visual text generation and editing by introducing explicit glyph, position, and masked-image conditions, together with OCR-aware design for accurate rendering \cite{anytext}. TextDiffuser-2 enhances automation and diversity by using language models for layout planning and line-level text rendering \cite{textdiffuser2}. These methods substantially improve text controllability, but they generally rely on dedicated training, extra control modules, or specialized conditioning pipelines \cite{textdiffuser,anytext,textdiffuser2}.

More recent work studies more challenging visual-text settings with stronger pretrained backbones. TextCrafter targets complex visual scenes with multiple text instances and improves rendering through text-insulation and text-oriented attention mechanisms \cite{textcrafter}. It also shows that quotation-related tokens can provide useful localization cues for quoted visual text regions, which motivates our use of opening-quotation-token attention for target-text localization. In contrast, FreeText is a training-free method for diffusion transformers: it decomposes text rendering into ``where to write'' and ``what to write'', localizes writing regions from endogenous attention, and strengthens glyph structure through spectral glyph injection \cite{freetext}. 

Compared with these visual text rendering methods, our goal is not to render text alone. We focus on a more specific and challenging setting where the intended text layout should remain readable while a newly generated object locally occludes part of the text region. Thus, glyph structure is used not as a standalone text-rendering solution, but as a stabilizing prior within an occlusion-aware generation framework.

\subsection{Occlusion-Aware and Structure-Preserving Editing}
Another closely related line concerns localized editing while preserving global structure. FreeFlux provides an especially relevant perspective for our work: it analyzes layer-specific roles in RoPE-based MMDiT models such as FLUX and designs task-specific K/V injection strategies for different editing regimes, including position-dependent editing and region-preserved editing \cite{freeflux}. This supports the view that stable editing on FLUX-like backbones requires careful selection of both intervention regions and intervention layers, rather than applying uniform manipulation throughout the network.

LaRender addresses occlusion more explicitly. It proposes a training-free latent rendering framework that controls occlusion relationships by using object ordering and transmittance-aware latent composition, enabling more precise front--back relations than prompt-only control \cite{larender}. While LaRender emphasizes general occlusion control in image generation, our setting is text-centric: the inserted object should occlude only a controlled subregion around the target text without destroying the surrounding typographic layout.

A related direction is structure locking. Anchor Token Matching, introduced in ISLock, preserves structural blueprints during training-free autoregressive image editing by dynamically aligning internal token correspondences with a reference image \cite{islock}. Although this method is developed for autoregressive visual generation rather than diffusion transformers, its core insight is relevant: localized editing benefits from preserving internal structural anchors instead of globally overriding the generation process.

Overall, our method lies at the intersection of visual text rendering, inference-time transformer control, and occlusion-aware editing. Similar to recent FLUX-oriented inference-time methods, we keep the pretrained backbone frozen and manipulate internal generation dynamics during sampling \cite{freeflux,freetext}. Similar to occlusion-aware editing methods, we explicitly care about local overlap and spatial front--back relationships \cite{larender}. However, unlike prior work, our target is \emph{text-preserving localized occlusion} on a pretrained FLUX.1-dev backbone, where the model must simultaneously maintain typographic layout and generate a plausible occluding object in a controlled text region.

\section{Methodology}

\subsection{Overview}
We propose a training-free restarted dual-stream inference framework built on a pretrained flow-matching backbone, e.g., FLUX.1-dev, for occluded text rendering. The key idea is to decouple \emph{text-layout preservation} from \emph{occluder insertion}. A \emph{Base Stream} is conditioned on a clean text-rendering prompt without the occluder and therefore provides a stable typographic reference, while an \emph{Edit Stream} is conditioned on the full occlusion prompt and is responsible for introducing the occluding object.

The proposed method takes a target text string, a clean \emph{Base prompt}, an occluder-aware \emph{Edit prompt}, shared initial noise, and a coarse text layout region for glyph-prior rasterization as inputs. The coarse text layout region is used only to place the glyph prior at an approximate target-text location. We refer to this spectral glyph-mask construction and injection module as \emph{spectral glyph-mask injection} (SGMI). The final occlusion mask is not manually specified; it is derived from token-conditioned text localization, glyph support, occluder-token attention, and anchor-aware selection.

\begin{figure*}[t]
    \centering
    \includegraphics[width=0.95\textwidth]{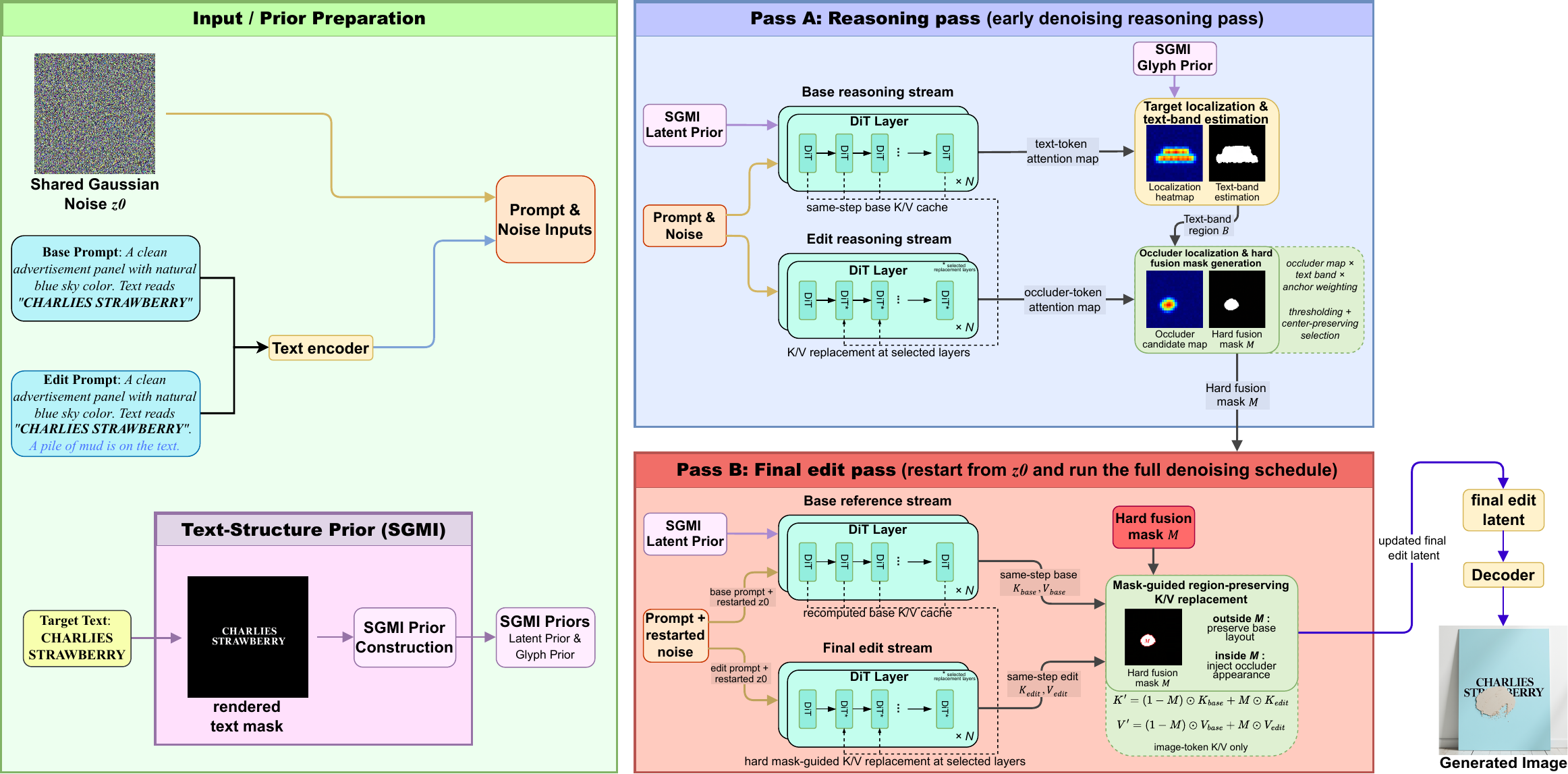}
    \caption{
    Overview of the proposed training-free occluded text rendering framework. The input stage prepares shared Gaussian noise, base/edit prompt embeddings, and SGMI-based glyph priors. In Pass A, the Base Reasoning Stream extracts a text-token localization map. The opening-quotation token can be used as a localization anchor for the quoted target text. This map is then combined with the glyph prior to estimate the text-band region $B$. Meanwhile, the Edit Reasoning Stream extracts an occluder-token attention map under same-step Base K/V guidance. The text-band region, occluder candidate map, and anchor weighting are combined to produce a hard fusion mask $M$. In Pass B, generation restarts from the same initial noise $z_0$. The Base Reference Stream recomputes same-step Base K/V caches, and the Final Edit Stream performs hard mask-guided image-token K/V replacement at selected layers, preserving the Base layout outside $M$ while injecting the occluder appearance inside $M$. The final denoised edit latent is decoded into the generated image.
    }
    \label{fig:architecture}
\end{figure*}

Unlike a single-pass generation pipeline, our method performs two sequential inference passes from the same initial Gaussian latent noise $z_0$. In the first pass, denoted as the \emph{reasoning pass}, the Base Stream extracts a text localization map and constructs a text-band region $B$ by combining token-conditioned attention with a glyph-based structural prior. This design follows prior observations that attention maps in text-to-image diffusion models provide useful spatial attribution for text-conditioned regions~\cite{daam,freetext}. For quoted visual text, we further adopt the observation from TextCrafter that quotation-related tokens can serve as effective attention anchors for the enclosed text region~\cite{textcrafter}. The Edit Stream then probes the occluding object under same-step Base K/V guidance and derives an anchor-aware candidate occlusion region. Fig.~\ref{fig:architecture} illustrates the overall pipeline of the proposed framework.

\subsection{Layer-wise Backbone Decomposition}
The underlying FLUX backbone follows recent transformer-based rectified-flow text-to-image designs, where text and image tokens interact through transformer attention rather than a U-Net denoiser~\cite{dit,scalingrf,flux}. It is composed of two structurally different stages: 19 \emph{double-stream blocks}, where text and image tokens are processed in parallel and interact through joint attention, followed by 38 \emph{single-stream blocks}, where the concatenated token sequence is processed jointly. This distinction is important for our method for two reasons. First, text-region localization is derived from token-conditioned image--text interaction signals before the final restarted edit pass. Second, our K/V intervention is applied through the attention operator and may affect both block types, but only at a sparse set of selected attention sites rather than over a continuous layer interval.

\subsection{Frequency-Filtered Glyph Prior}
To stabilize typography during generation, we incorporate a glyph-based structural prior into the Base Stream. Given the target text and a coarse text layout region, we first rasterize a binary glyph mask in image space. This region is used only to place the glyph prior at an approximate target-text location, rather than to directly specify the final occlusion mask. This mask is then converted into two latent forms.

First, the rasterized mask is resized to the latent resolution and flattened into a spatial gate,
\begin{equation}
    G \in \mathbb{R}^{L_{\mathrm{img}} \times 1},
\end{equation}
which restricts the prior to the target text region.

Second, the rendered glyph image is encoded by the pretrained autoencoder and passed through a frequency-domain filtering operation, yielding a packed latent prior
\begin{equation}
    P_{\mathrm{glyph}} \in \mathbb{R}^{L_{\mathrm{img}} \times C}.
\end{equation}

The glyph prior is injected only during an early-to-mid denoising window:
\begin{equation}
\begin{gathered}
    x^{\mathrm{base}}_{s,\mathrm{in}} =
    x^{\mathrm{base}}_{s}
    + \lambda_{\mathrm{glyph}} \left(P_{\mathrm{glyph}} \odot G\right), \\
    p_s \in [\alpha, \beta].
\end{gathered}
\end{equation}
Here, $\lambda_{\mathrm{glyph}}$ is the injection strength, $[\alpha,\beta]$ defines the active progress range, and $p_s$ denotes the normalized denoising progress at step $s$. Its role is to preserve coarse glyph structure in the latent trajectory rather than to synthesize the final appearance of the text.

\subsection{Pass A: Base-side Text Localization and Text-Band Construction}
The Base Stream is conditioned on a clean prompt that contains the scene description and the target text, but excludes the occluding object. Its role is to establish a clean typographic trajectory and provide same-step K/V references for the Edit Stream during the reasoning pass. These K/V references are used only within the reasoning pass; after restart, the Final Edit Pass recomputes the Base Stream references.

At denoising step $s$, the Base Stream predicts a flow direction
\begin{equation}
    v^{\mathrm{base}}_s =
    f_\theta\!\left(x^{\mathrm{base}}_s, c_{\mathrm{base}}, s\right),
\end{equation}
and updates the latent by
\begin{equation}
    x^{\mathrm{base}}_{s+1}
    =
    x^{\mathrm{base}}_s + \Delta t_s v^{\mathrm{base}}_s.
\end{equation}
For each attention site $l$, we cache the same-step key and value tensors:
\begin{equation}
    \left(K^{\mathrm{base}}_{s,l}, V^{\mathrm{base}}_{s,l}\right).
\end{equation}

To localize the target text, we follow prior cross-attention attribution analysis and training-free text localization methods, which show that token-conditioned attention maps can reveal spatial regions associated with specific text prompts~\cite{daam,freetext}. We extract image-to-text attention responses associated with a set of text localization indices $\mathcal{Q}$. In practice, $\mathcal{Q}$ can include either the tokenized target phrase or the opening-quotation token preceding the target text. The latter is particularly useful when the target phrase is decomposed into multiple sub-tokens. Following the quotation-token observation used in TextCrafter~\cite{textcrafter}, the preceding quotation mark provides a compact attention anchor for the visual text enclosed by quotation marks. We denote the aggregated text localization map at the reasoning cutoff step $s_r$ as
\begin{equation}
    A^{\mathrm{text}}_{s_r}
    =
    \Phi_{\mathrm{loc}}(\mathcal{Q}).
\end{equation}
Here, $\Phi_{\mathrm{loc}}(\cdot)$ denotes the aggregation of image-token responses associated with the selected text-token indices from the joint attention matrix of FLUX transformer blocks.

Instead of using only a single center point, we combine this localization map with the glyph-derived structural prior to estimate a soft text-band region:
\begin{equation}
    B \in [0,1]^{L_{\mathrm{img}}}.
\end{equation}
The text band provides both a center estimate and a spatial support region for subsequent occluder localization. This design makes the occlusion mask depend on the generated text layout rather than on a manually specified insertion point.

\subsection{Pass A: Edit-side Object Probing and Anchor-Aware Mask Derivation}
The Edit Stream is conditioned on the full edit prompt, which augments the Base prompt with an occluding object description. During the reasoning pass, it starts from the same initial noise $z_0$ as the Base Stream. At selected transformer blocks, its image-token K/V tensors are replaced by the same-step Base Stream K/V tensors:
\begin{equation}
\begin{aligned}
    K^{\mathrm{edit,img}}_{s,l} &\leftarrow K^{\mathrm{base,img}}_{s,l}, \\
    V^{\mathrm{edit,img}}_{s,l} &\leftarrow V^{\mathrm{base,img}}_{s,l}, \\
    l &\in \mathcal{S}.
\end{aligned}
\end{equation}
This reasoning-mode replacement is not the final occlusion fusion. Instead, it stabilizes the Edit Stream under the clean Base layout while allowing the occluder semantics to emerge.

At the reasoning cutoff step $s_r$, we similarly use token-conditioned attention as a spatial attribution signal to extract an occluder-token localization map from the Edit Stream~\cite{daam}. Let $\mathcal{O}$ denote the token indices corresponding to the occluding object. We write
\begin{equation}
    A^{\mathrm{obj}}_{s_r}
    =
    \Phi_{\mathrm{loc}}(\mathcal{O}).
\end{equation}

The occluder map is then constrained by the text-band region $B$ and a center-aware anchor mask $W_{\mathrm{anchor}}$. In our implementation, the anchor is used as a weighted spatial prior rather than a strict multiplication mask:
\begin{equation}
    \widetilde{A}^{\mathrm{obj}}
    =
    A^{\mathrm{obj}}_{s_r}
    \odot B
    \odot
    \left((1-\rho) + \rho W_{\mathrm{anchor}}\right),
\end{equation}
where $\rho$ controls the strength of the anchor weighting.

After smoothing and thresholding, we select the connected component closest to the text-band center. This produces a soft candidate region $\widetilde{M} \in [0,1]^{L_{\mathrm{img}}}$. We then apply hard thresholding and dilation to obtain the final hard fusion mask
\begin{equation}
    M \in \{0,1\}^{L_{\mathrm{img}}}.
\end{equation}
In the implementation, $M$ is broadcast to the head and channel dimensions of the image-token K/V tensors.
If the semantic candidate is unreliable or nearly empty, we fall back to an anchor-centered text-band mask. As a result, the final fusion mask is determined jointly by text localization, glyph support, occluder-token attention, and center-aware selection.

\subsection{Pass B: Restarted Final Edit Pass with Hard Mask-Guided Image-Token K/V Replacement}
After obtaining the hard fusion mask $M$, we restart generation from the same initial noise $z_0$ and run the full denoising schedule again. In this restarted pass, the Base Stream is recomputed at every step to refresh same-step K/V references, while the Edit Stream performs hard mask-guided image-token K/V replacement at selected attention sites.

A key implementation detail is that the replacement is applied only to image tokens. The text-token K/V tensors of the Edit Stream are left unchanged, so the edit prompt remains the semantic condition for final generation. Let
\begin{equation}
    K^{\mathrm{base,img}}_{s,l},\; V^{\mathrm{base,img}}_{s,l}
\end{equation}
and
\begin{equation}
    K^{\mathrm{edit,img}}_{s,l},\; V^{\mathrm{edit,img}}_{s,l}
\end{equation}
denote the image-token portions of the Base and Edit Stream tensors at denoising step $s$ and attention site $l$. For $l \in \mathcal{S}$, we compute
\begin{equation}
\begin{gathered}
    K^{\prime\,\mathrm{img}}_{s,l}
    =
    (1-M)\odot K^{\mathrm{base,img}}_{s,l}
    +
    M \odot K^{\mathrm{edit,img}}_{s,l},
\\
    V^{\prime\,\mathrm{img}}_{s,l}
    =
    (1-M)\odot V^{\mathrm{base,img}}_{s,l}
    +
    M \odot V^{\mathrm{edit,img}}_{s,l}.
\end{gathered}
\end{equation}
Since $M$ is a binary hard mask, this masked mixture acts as a region-wise K/V replacement: outside $M$, the Edit Stream inherits the Base Stream K/V to preserve the clean text layout; inside $M$, it keeps the Edit Stream K/V to introduce the occluder appearance.

The text-token part remains unchanged:
\begin{equation}
\begin{gathered}
    K^{\prime\,\mathrm{text}}_{s,l}
    =
    K^{\mathrm{edit,text}}_{s,l},
\\
    V^{\prime\,\mathrm{text}}_{s,l}
    =
    V^{\mathrm{edit,text}}_{s,l}.
\end{gathered}
\end{equation}

This operation affects the attention computation inside the Final Edit Stream rather than acting as a post-processing step. Therefore, the K/V replacement gradually changes the denoising trajectory of the Edit Stream. After all denoising steps are completed, the final denoised edited latent is passed to the decoder.

\subsection{Final Decoding}
The hard mask-guided K/V replacement does not directly output an image. Instead, it modifies the attention computation within the Final Edit Stream and thereby changes the latent trajectory. After the full denoising schedule is completed, the final latent from the Edit Stream is decoded by the pretrained autoencoder:
\begin{equation}
    I_{\mathrm{out}}
    =
    \mathrm{AE}_{\mathrm{dec}}\!\left(x^{\mathrm{edit}}_{N}\right).
\end{equation}
The resulting image preserves the intended typography layout from the Base Stream while inserting the occluding object in the mask-guided region.

\subsection{Implementation-specific Settings}
Although the method is defined using generic variables, our current implementation uses $N=28$ denoising transitions, reasoning cutoff step $s_r=7$, and glyph-prior active range $[\alpha,\beta]=[0.1,0.4]$. For sparse K/V replacement, we adopt the position-relevant attention sites identified by FreeFlux~\cite{freeflux}, which performs a layer-wise probing analysis of RoPE-based MMDiT models and reports that the layers
\begin{equation}
    \mathcal{S}
    =
    \{1,2,4,26,30,54,55\}
\end{equation}
are highly position-dependent for object addition. Since our occluded text rendering task also requires inserting an occluding object at a spatially constrained text region, we use this set as the sparse replacement sites in both the reasoning and final edit passes.

The Final Edit Pass is restarted from the same shared initial noise and applies hard mask-guided image-token K/V replacement throughout the denoising schedule at these selected sites. These values are implementation choices rather than intrinsic constraints of the proposed framework.

\section{Experiments}

\subsection{Experimental Setup}

\paragraph{Task Setting}
We evaluate the proposed method on occluded text rendering, where the generated image is expected to satisfy three requirements simultaneously: 
(1) the target text should remain recognizable, 
(2) the occluding object should spatially overlap the intended text region, and 
(3) the overall composition should remain visually coherent. 
This setting is more challenging than ordinary text-to-image generation because the model must not only render readable text, but also establish a physically plausible front--back relationship between the text and the occluding object.

\paragraph{Compared Methods}
We compare our method with both general text-to-image models and occlusion-aware generation baselines:
\begin{itemize}
    \item \textbf{SDXL~1.0}~\cite{sdxl}: a widely used latent diffusion baseline with strong general text-to-image generation capability.
    \item \textbf{FLUX.1-dev}~\cite{flux}: the rectified-flow transformer backbone used by our method, serving as the most direct baseline.
    \item \textbf{LaRender-GLIGEN}~\cite{larender,gligen}: a training-free occlusion-control baseline that explicitly targets object occlusion relationships.
    \item \textbf{HiDream-I1-Full}~\cite{hidream}: a large-scale open-source image generation model with strong visual generation quality.
    \item \textbf{Qwen-Image-2512}~\cite{qwenimage}: a recent image generation model with strong text rendering and image editing capabilities.
    \item \textbf{Ours}: the proposed restarted dual-stream framework with glyph priors, text-band reasoning, and hard mask-guided image-token K/V replacement.
\end{itemize}

SDXL and FLUX.1-dev are used to evaluate whether general text-to-image models can directly solve the occluded text rendering task. LaRender-GLIGEN is included because it explicitly addresses occlusion control. HiDream-I1-Full and Qwen-Image-2512 are included as stronger recent generation models; although they often produce visually appealing results, they also require substantially higher GPU memory usage during inference. This comparison allows us to evaluate both generation quality and computational practicality.

\paragraph{Evaluation Dataset}
We evaluate all methods on 64 generated samples, consisting of 8 occluded-text prompts across four scenario categories and 8 seeds for each prompt. The scenario categories are as follows: printed text on T-shirts, poster text occluded by a vinyl record, text on an airship occluded by a tree or branch, and graffiti text on a wall occluded by a tree trunk. Each scenario specifies a target text string, an occluding object, and an approximate text region for evaluation. For baseline methods, we use a single prompt describing the final occluded scene. For our method, we use a clean \emph{Base prompt} without the occluder and an occluder-aware \emph{Edit prompt}.

\paragraph{Implementation Details and VRAM Usage}
For all baselines, we use the same set of target scene descriptions and generate images at $1024 \times 1024$ resolution. When applicable, we use the same random seeds and report results over the same 64 generated samples. For methods that require additional layout or grounding inputs, such as LaRender-GLIGEN, we provide the corresponding occluder description and use the same approximate target text region when applicable. For our method, we use $N=28$ denoising transitions, reasoning cutoff step $s_r=7$, glyph-prior active range $[0.1,0.4]$, and sparse replacement sites $\{1,2,4,26,30,54,55\}$. Peak VRAM is measured on a single NVIDIA RTX PRO 6000 GPU with batch size 1 and bfloat16/float16 inference where supported. As shown in Table~\ref{tab:vram_comparison}, our method introduces moderate overhead compared with pure FLUX.1-dev due to the restarted dual-stream inference, but remains substantially lighter than HiDream-I1-Full and Qwen-Image-2512.

\begin{table}[t]
    \centering
    \caption{Approximate peak VRAM usage of different methods under our evaluation setting.}
    \label{tab:vram_comparison}
    \begin{tabular}{lc}
        \toprule
        Method & Peak VRAM Usage \\
        \midrule
        SDXL~1.0 & 17 GB \\
        LaRender-GLIGEN & 18 GB \\
        FLUX.1-dev & 25 GB \\
        Ours & 29 GB \\
        Qwen-Image-2512 & 55 GB \\
        HiDream-I1-Full & 66 GB \\
        \bottomrule
    \end{tabular}
\end{table}

\paragraph{Evaluation Metrics}
We report three quantitative metrics tailored to occluded text rendering:
\begin{itemize}
    \item \textbf{Text similarity} ($\mathrm{text\_sim}$): We first recognize the generated text using an OCR engine, and then compute the normalized edit similarity between the recognized text and the target text. This metric evaluates whether the target text remains readable after occlusion.
    \item \textbf{Occlusion alignment} ($\mathrm{occ\_align}$): We estimate the spatial overlap between the detected occluder region and the target text region. To penalize both insufficient overlap and overly large occlusion, we use a balanced overlap score:
    \begin{equation}
        \mathrm{occ\_align}
        =
        \sqrt{
        \frac{|R_{\mathrm{occ}} \cap R_{\mathrm{text}}|}{|R_{\mathrm{occ}}|}
        \cdot
        \frac{|R_{\mathrm{occ}} \cap R_{\mathrm{text}}|}{|R_{\mathrm{text}}|}
        }.
    \end{equation}
    If the occluder is not detected, we set $\mathrm{occ\_align}=0$ for that sample.
    \item \textbf{Occluder detection rate} ($\mathrm{detect\_rate}$): We use an open-set object detector to verify whether the intended occluding object is successfully generated. This metric measures the object-generation reliability of each method.
\end{itemize}
In practice, OCR is performed using EasyOCR~\cite{easyocr}, and occluder detection is performed with GroundingDINO~\cite{groundingdino}. EasyOCR is a general OCR toolkit for extracting text from images, while GroundingDINO supports open-set detection using category names or referring expressions. 

\subsection{Comparison with Existing Methods}

\begin{figure*}[t]
    \centering
    \includegraphics[width=\textwidth]{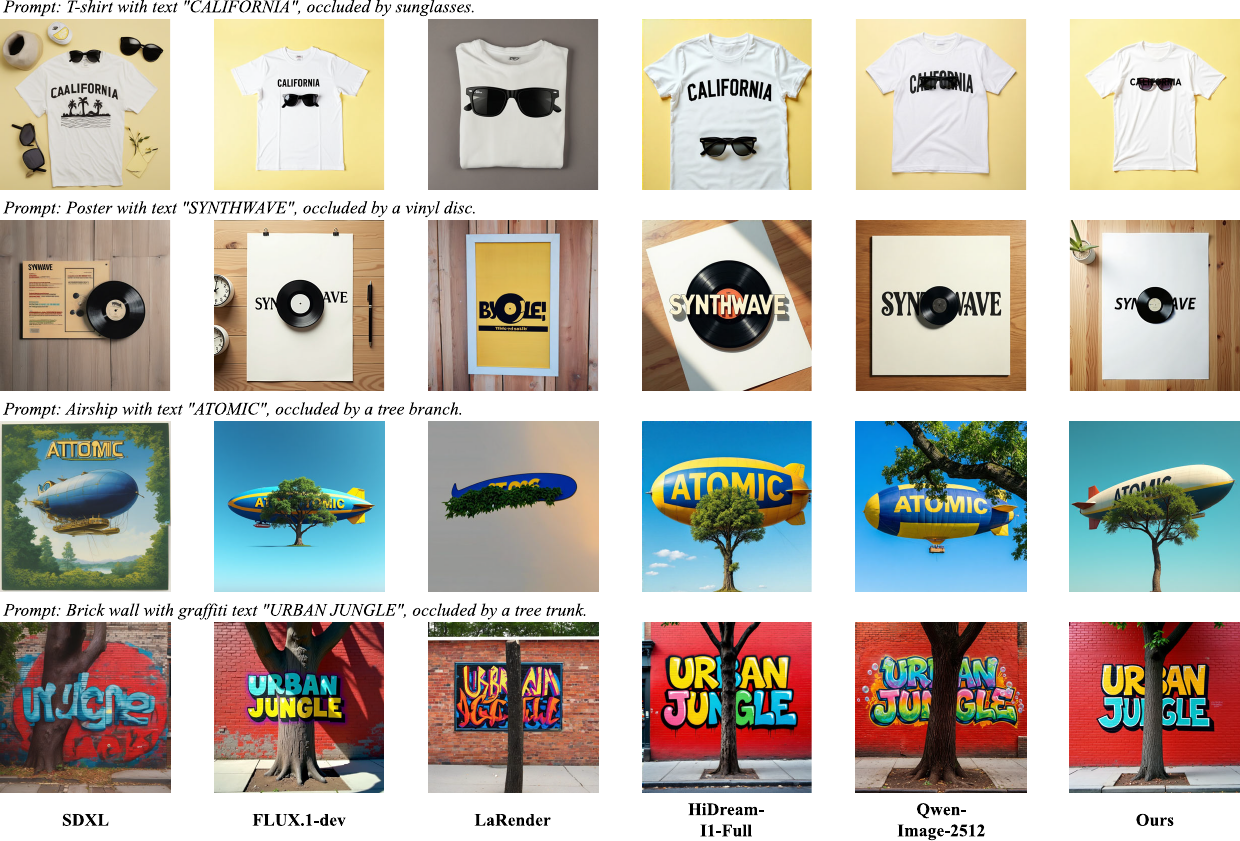}
    \caption{
    Qualitative comparison on representative occluded text rendering cases. Each row corresponds to one scene, and each column corresponds to one method. Compared with general text-to-image models and occlusion-control baselines, our method better preserves the target typography while placing the occluding object on the intended text region.
    }
    \label{fig:main_qualitative_comparison}
\end{figure*}

\paragraph{Qualitative Comparison}
Fig.~\ref{fig:main_qualitative_comparison} shows representative qualitative results. SDXL~1.0 often fails to preserve the target text faithfully and may generate incorrect or distorted typography. FLUX.1-dev improves the overall visual quality, but the occluding object may drift away from the intended text region or fail to establish a convincing front--back relationship. LaRender-GLIGEN provides stronger occlusion control, but its text rendering quality is less stable in typography-sensitive scenes. HiDream-I1-Full and Qwen-Image-2512 produce visually strong images and sometimes achieve good object placement, but their text rendering and occlusion relationships remain inconsistent across different scenes. In contrast, our method better preserves the target text layout while placing the occluder on the intended text region.

\paragraph{Quantitative Comparison}
Table~\ref{tab:main_quantitative_comparison} reports the quantitative comparison on 64 generated images. Our method achieves the highest text similarity among all compared methods and obtains competitive occlusion alignment while maintaining a perfect occluder detection rate. Although LaRender-GLIGEN obtains a higher $\mathrm{occ\_align}$ score, its text similarity is substantially lower, indicating that explicit occlusion control alone does not guarantee readable typography. In contrast, our method provides a better balance between text readability and object-on-text placement.

Compared with pure FLUX.1-dev, our method improves $\mathrm{text\_sim}$ from 0.6933 to 0.8895 and $\mathrm{occ\_align}$ from 0.3357 to 0.4162. This confirms that simply prompting FLUX.1-dev with an occluded-text description is insufficient for stable spatial occlusion. SDXL~1.0 shows weaker text rendering and lower occlusion alignment, while HiDream-I1-Full and Qwen-Image-2512 show competitive visual quality but require substantially higher VRAM and remain below our method on text similarity and slightly below our method on occlusion alignment.

\begin{table}[t]
    \centering
    \footnotesize
    \setlength{\tabcolsep}{3.2pt}
    \caption{
    Quantitative comparison on 64 occluded-text rendering samples. Higher values are better for all metrics.
    }
    \label{tab:main_quantitative_comparison}
    \begin{tabular}{lccc}
        \toprule
        Method & $\mathrm{text\_sim}\uparrow$ & $\mathrm{occ\_align}\uparrow$ & $\mathrm{detect\_rate}\uparrow$ \\
        \midrule
        SDXL~1.0 & 0.5716 & 0.2709 & 0.8125 \\
        FLUX.1-dev & 0.6933 & 0.3357 & 0.9375 \\
        LaRender-GLIGEN & 0.2117 & \textbf{0.5140} & \textbf{1.0000} \\
        HiDream-I1-Full & 0.6904 & 0.3243 & 0.9063 \\
        Qwen-Image-2512 & 0.6961 & 0.4067 & \textbf{1.0000} \\
        Ours & \textbf{0.8895} & 0.4162 & \textbf{1.0000} \\
        \bottomrule
    \end{tabular}
\end{table}

\subsection{Ablation Study}

We further analyze the contribution of the main components in the proposed framework using the same scene set, target text strings, approximate text regions, and random seeds as the main comparison. Unlike the main comparison, the ablation study is designed to isolate individual modules rather than to compare complete occluded-text generation systems. Therefore, text-only variants such as \emph{FLUX} and \emph{FLUX + SGMI} are evaluated with clean prompts that exclude the occluding object, so that we can measure their ability to render the base typography. For variants that explicitly introduce an occluder, we additionally report \emph{occlusion alignment} ($\mathrm{occ\_align}$) and \emph{occluder detection rate} ($\mathrm{detect\_rate}$).

\paragraph{Ablation Variants}
We consider four representative variants:
\begin{itemize}
    \item \textbf{FLUX}: a text-only vanilla FLUX.1-dev variant conditioned on the clean base prompt, without the occluder, SGMI, or the proposed occlusion mechanism. This variant is used to evaluate the native typography-rendering ability of the backbone.
    \item \textbf{FLUX + SGMI}: a text-only variant that augments FLUX.1-dev with the SGMI-based text-structure prior, while still excluding the occluder. This variant isolates the effect of the glyph prior on text rendering.
    \item \textbf{FLUX + Stacking}: a simplified occlusion variant that introduces the occluder without the full SGMI-enhanced framework.
    \item \textbf{Full Model}: the complete proposed method with both SGMI and the occlusion mechanism.
\end{itemize}

\paragraph{Quantitative Ablation}
Table~\ref{tab:ablation} summarizes the quantitative ablation results. In the text-only setting, adding SGMI increases $\mathrm{text\_sim}$ from 0.6845 to 0.8985, showing that the text-structure prior is highly effective for stabilizing base typography. Since both \emph{FLUX} and \emph{FLUX + SGMI} use clean prompts without the occluder, $\mathrm{occ\_align}$ and $\mathrm{detect\_rate}$ are not applicable to these two variants.

Compared with the text-only variants, the \emph{FLUX + Stacking} variant introduces the occluder but does not include the complete SGMI-enhanced reasoning and hard mask-guided K/V replacement mechanism. Although it enables occluder generation, the improvement in text readability and occlusion alignment remains limited. This suggests that simply adding an occluder branch is insufficient for reliably preserving typography while controlling object-on-text placement.

\begin{table}[t]
    \centering
    \scriptsize
    \setlength{\tabcolsep}{2.5pt}
    \caption{
    Ablation study on the same 64 samples as the main comparison. Text-only variants use clean prompts without the occluder. Higher values are better for applicable metrics, and ``--'' denotes non-applicable metrics.
    }
    \label{tab:ablation}
    \begin{tabular}{@{}lccc@{}}
    \toprule
    Variant & \makecell{$\mathrm{text\_sim}$ $\uparrow$} 
            & \makecell{$\mathrm{occ\_align}$ $\uparrow$} 
            & \makecell{$\mathrm{detect\_rate}$ $\uparrow$} \\
    \midrule
    FLUX (text-only) & 0.6845 & -- & -- \\
    FLUX+SGMI (text-only) & \textbf{0.8985} & -- & -- \\
    FLUX+Stacking & 0.7084 & 0.3355 & 0.9375 \\
    Full Model & 0.8895 & \textbf{0.4162} & \textbf{1.0000} \\
    \bottomrule
    \end{tabular}
\end{table}

\begin{figure}[t]
    \centering
    \includegraphics[width=0.92\columnwidth]{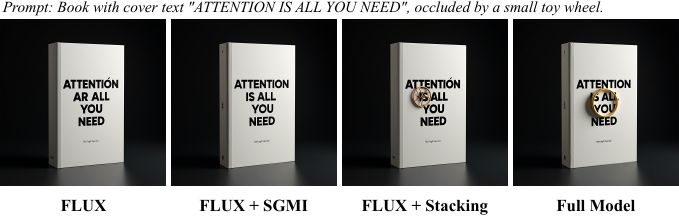}
    \caption{
    Qualitative ablation. SGMI improves text structure, while the full model balances text preservation and occlusion placement.
    }
    \label{fig:ablation_qualitative}
\end{figure}

\paragraph{Qualitative Ablation}
Fig.~\ref{fig:ablation_qualitative} provides a representative qualitative ablation example. The text-only FLUX variant reflects the native typography-rendering ability of the backbone under the clean base prompt, but its text structure remains unstable or inaccurate. Adding SGMI substantially improves typography quality and corrects the target text structure. The \emph{FLUX + Stacking} variant can introduce the occluder, but the local composition and text preservation remain weaker. The \emph{Full Model} produces the most balanced result, achieving both readable text and a plausible occlusion relationship. This qualitative evidence is consistent with the quantitative trends in Table~\ref{tab:ablation}.

\section{Conclusion and Limitations}

\subsection{Conclusion}
We presented a training-free framework for occluded text rendering with a pretrained FLUX.1-dev backbone. By combining glyph-prior regularization, dual-stream reasoning, text-band localization, and hard mask-guided image-token K/V replacement, our method substantially improves text readability while achieving competitive occluder placement. Experiments show that the proposed framework produces more stable object-on-text compositions without requiring model fine-tuning.

\subsection{Limitations}
Our method still depends on the scene understanding and generative capability of the underlying FLUX backbone. When the occluding object is transparent, translucent, reflective, or structurally thin, the generated occlusion may be physically inaccurate or visually unstable. In addition, inaccurate text-band localization may lead to suboptimal mask estimation in highly cluttered scenes.

\section*{Acknowledgment}

We gratefully acknowledge Beijing Zhongke Dayang Technology Development Inc.\ for providing computational support for this research. We also thank Dr. Linxiao Niu for his valuable guidance and helpful discussions throughout this work.

\bibliographystyle{IEEEtran}

\end{document}